\documentclass{article}
\usepackage{spconf,amsmath,graphicx}

\let\Item\item

\renewcommand\enddescription{\endlist\global\let\item\Item}

\def\dialogue#1{  
  \begin{table}[H]
    \setlength{\tabcolsep}{3pt}
    \begin{tabular}{l p{6cm} }
    #1
    \end{tabular}
  \end{table}
  \vspace{-0.5em}
\noindent\ignorespaces}

\def\user#1{
\textbf{User:} & \textit{#1} \\
}
\def\system#1{
\textbf{System:} & \textit{#1} \\
}

\def\centeredutterance#1{
 \vspace{-0.70em}
  \begin{center}
    	\textit{#1}
  \end{center}
  \vspace{-0.5em}
\noindent\ignorespaces
}

\usepackage{times}
\usepackage{latexsym}

\usepackage{url}

\usepackage{float}
\usepackage{multibib}
\newcites{languageresource}{Language Resources}
\usepackage{graphicx}
\usepackage{amsmath}
\usepackage{epstopdf}
\usepackage[utf8]{inputenc}

\usepackage{enumitem}
\usepackage{url}
\usepackage[normalem]{ulem} % for \markoverwith

\usepackage{todonotes}
\presetkeys{todonotes}{inline}{}
\presetkeys{todonotes}{prepend}{}
\presetkeys{todonotes}{caption=TODO}{}

\definecolor{darkgreen}{rgb}{0.0, 0.5, 0.0}

\def\ODdel#1{\bgroup\markoverwith{\textcolor{darkgreen}{\rule[0.5ex]{2pt}{1pt}}}\ULon{#1}}
\def\fjd#1{\bgroup\markoverwith{\textcolor{red}{\rule[0.5ex]{2pt}{1pt}}}\ULon{#1}}

\def\newcite#1{\cite{#1}}

\title{Denotation Extraction for Interactive Learning in Dialogue Systems}
\name{Miroslav Vodolán, Filip Jurčíček}
\address{
	Charles University in Prague, Faculty of Mathematics and Physics  \\
  Institute of Formal and Applied Linguistics \\
  Malostranské náměstí 25, 11800 Praha 1, Czech Republic \\
  {\tt \{vodolan, jurcicek\}@ufal.mff.cuni.cz}
  }
\begin{document}
%\ninept
%
\maketitle
\begin{abstract}
This paper presents a novel task using real user data obtained in human-machine conversation.
%to improve quality of dialogue systems. 
The task concerns with denotation extraction from answer hints collected interactively in a dialogue. 
%\OD{Asi bych zacal tim, k cemu to je, pak co to je a potom mluvil o real user data... novel task enabling SDS to learn from users. It is concerned with denotation... Nekde na konci toho summary "We use real user data... in our experiments."}
%\MV{zmeny v abstraktu nejsou povolene}
%In contrast, the answer hints can be obtained quite easily in interactive learning setup as shown in~\newcite{vodolan_rewochat2016}. 
%Our previous work shown that the interactive collection of answer hints in a dialogue is feasible.
% and suggested that it has a potential to improve the dialogue systems. 
%The data we are focused on are user's answer hints in a form of natural language (English). 
%
The task is motivated by the need for large amounts of training data for question answering dialogue system development, where the data is often expensive and hard to collect. 
%
%The use of answer hints has a potential to ease collection of training data. 
%in interactive learning setup as shown in~\newcite{vodolan_rewochat2016}. 
%This work focuses on processing the answer hints into denotations during a process called denotation extraction.
%
%The denotations have already been used in training natural language understanding components in question answering systems. 
%Being able to collect annotation interactively, one could improve, for example, natural understanding components on-line.
Being able to collect denotation interactively and directly from users, one could improve, for example, natural understanding components on-line and ease the collection of the training data.

%One way of using the answer hint information is by extracting denotations which are useful for several dialogue system components as was already shown in related work. 

%The extracted denotations can be later used to enhance knowledge of a dialogue system and consequently improve the overall system performance.
This paper also presents introductory results of evaluation of several denotation extraction models including attention-based neural network approaches.
\end{abstract}
\section{Introduction}
\label{sec:introduction}
% s tim, jak se rozsiruje oblibenost dialogovych systemu a mnozstvi dat, ktere mohou svym uzivatelum potencialne nabizet take narusta poptavka po systemech, ktere je mozne nasazovat rychle na ruzne domeny dat.
The increasing popularity of dialogue systems and the rapidly growing amount of available information (factoid general data such as Wikidata, domain-specific data such as transportation schedules, etc.), demands dialogue systems which can be quickly adapted to new domains.

% tato potreba dala vzniknout nekolika industry technologiim, ktere umoznuji rychlou tvorbu dialogovych systemu. (jako Facebook Wit.ai nebo IBM Cognea)
This demand has motivated several industry technologies providing tools for quick development of dialogue systems, such as \textit{IBM Watson Dialogue Service}\footnote{https://www.ibm.com/watson/developercloud/dialog.html} and \textit{Wit.ai}\footnote{https://wit.ai}.
%The basic principle of these technologies is letting a domain expert to design the dialogue tree, i.e., a description of possibilities which could appear throughout the conversation. 
%The tree is then continuously extended and corrected by the domain expert based on feedback from users of the dialogue system. 
%
The core principle of these technologies is based on manual work of domain experts.
% kvuli tomuto principu byvaji vysledne dialogove systemy obvykle jednoduche a zamerene na jednu domenu
This principle of handcrafting usually causes the resulting dialogue systems to be quite simple, focused on a single domain (separate systems for pizza ordering, restaurant bookings, transportation information, etc.), and they have troubles when users are trying to use out-of-domain concepts as noted in~\cite{bordes2016learning,pappu2013predicting}.
A typical problem of such systems is that they are hard to maintain in the long term as language usage evolves (e.g. how users refer to concepts in a domain) and a domain itself changes (e.g. new concepts are added).  
To sum up, a principled approach to dialogue system development must be able to adapt in the long term.
%
%\fj{to chce predelat, ten problem out of domain concepts nenti az takovy z duvodu ze to je handcrafted, ale ze se tyto veci v case vyviji, ze to vlastne ani nejde handcraftnout. ze nove pojmy toho stejneho se objevuji a zanikaji, a naucit se to musime od samotnych uzivatelu, oni maji tu znalost o zmene/rozsireni jazyka. OD: Souhlas. Tady je potreba explicitne napsat, ze je nutne vyvijet data-driven models, abys o nich pak mohl mluvit v dalsim odstavci}
%\MV{na vysvetleni a priklad ze se domeny stale vyviji zatim neni misto. O data driven pisu v nasledujicim odstavci}

% Abychom mohli pomoci supervised learning natrenovat komplexni model, potrebujeme velke mnozstvi anotovanych dat, ktera vsak nemusi byt dostupna.
Models with more complexity, covering many domains, could be developed on the basis of supervised machine learning techniques, as shown by recent end-to-end dialogue systems~\cite{wen2016end2end}. 
%\fj{Citace prace poti kterym se vyhrazujes, nejake E2E sytemy, mozna nekete po castech ucene.}
Such data-driven models require large amounts of labeled training data and many tricks to make them work, e.g. training components of the system separately with different  training data~\cite{wen2017latent}.
The issue with those models is in obtaining sufficient amounts of labeled training data, which is often not feasible for a non-trivial number of domains, especially when the data must be collected repetitively.%\OD{repeatedly}

Some research focus on reinforcement learning (RL) to alleviate the problem. 
The RL methods allows for on-line adaption of dialogue policies and learning from user feedback~\cite{gavsic2017dialogue}.
%\fj{citace steve young, blaise, gasic}. 
%\MV{ktere citace mas na mysli?x}
%In some case, a joint learning of dialogue state tracking and dialogue policy is possible \fj{citace, fj NABC journal}
While a lot of progress has been made,
%in usage of the RL in the context spoken dialogue systems, 
the proposed methods are not very efficient yet.
In part, it is caused by learning only from feedback in the form of a delayed numerical reward.
Consequently, these methods only learn and adapt a model of communication, e.g. what response (often a handcrafted action) to use given the facts the dialogue system has access to.
This severely limits their ability to adapt in the long term.
%\fj{This severally limits their ability to adapt in long term as the language evolves (e.g. new ways how to express existing facts) or the world changes (e.g. facts changes or new facts become true).}
%
%One would expect trainable dialogue systems  

%
%\fj{They do not learn facts about the world and how these facts are realized in surface forms.}
%The proposed methods do not attempt to learn from explicit feedback in conversation such as "No, you are wrong."
%They do not learn alternative names to entities, do not learn new spoken expression representing facts, cannot learn to communicate about new facts in a knowledge base.  
%They do not adapt to 

% Alternativnim pristupem je interaktivni uceni (vodolan lrec, Weston-Bordes ILRC), ktere umoznuje ziskat data vhodna pro trenink primo z dialogu od uzivatelu.
An alternative approach called \emph{interactive learning}~\cite{azaria2016instructable,li2016learning,weston2017interactivelearning,vodolan_rewochat2016} aims at obtaining factual information directly from conversations with real users and eventually use such information in real-time to improve skills of a dialogue system. 
Therefore, it has the potential to allow developing dialogue systems with less demand on domain experts and labeled training data.
Interactive learning represents a great step towards adaptability of dialogue systems.

The work of~\newcite{azaria2016instructable} uses interactive learning to teach a personal agent new skills by combining old ones. 
Other work shows the possibility to use the interactive learning for improving language understanding of a system~\cite{li2016learning,weston2017interactivelearning}.
The authors  propose a user simulator which can be questioned by the system about simple facts. 
This setup neglects the complexity of the natural language (due to simple templates that the simulator uses), which is an issue for real dialogue system deployment.
%The issue with this setup is that it is neglecting the complexity of the natural language (due to simple templates that the simulator uses).
%
Interactive learning in connection with natural language is researched in~\cite{vodolan_rewochat2016}. 
The authors collected a large set of natural dialogues, called the Question Dialogue Dataset (QDD), to be used for experiments with interactive learning. 
They distinguish several kinds of information which an interactive system can obtain during a dialogue with a user. 
The most basic kind of information is an \emph{answer hint}, which is obtained by asking a human user a factual question of system's interest. 
%The dataset is called the Question Dialogue Dataset (QDD).

%The work distinguishes three kinds of information which an interactive system can obtain during a dialogue with a user. 
%First, an \emph{answer hint} which is obtained by asking a human user a factual question of system's interest. 
%Second, a \emph{language information} that helps the system improve its language understanding abilities.
%Finally, a \emph{deep semantic concept understanding} which can help the system improve its reasoning process.

This paper proposes a task which aims to use answer hints to derive question denotations~\cite{Berant2013} (see~Section~\ref{sec:task} for details). 
%A denotation is a form of meaning representation in Question-Answering systems. 
For example, manually annotated denotations are used to train natural language understanding (NLU) components~\cite{Berant2013,liang2016learning,liang2016neural}.
The denotation use itself is outside of the scope of this work.
However, the ability to obtain the denotations automatically in real-time potentially enables dialogue systems to improve their NLU interactively while using, for example, the referred techniques above.
Another possibility is to use denotations to enrich a system's knowledge base (see~Section~\ref{sec:task} for details).
%A system may ask questions for which it does not know answers. 
A system may ask questions about topics with poor coverage. 
Then, the answers (denotations) can be used to derive new facts.
This relates to the work on knowledge base population~\cite{ji2011knowledge,mcnamee2009}.
%\fj{citace}  (https://arxiv.org/abs/1506.00301 ).
%\fj{POZNAMKA: Rozsireni tohoto clanky by mohlo byt ze by jsme pouzili tento denotation extraction algoritmu v live systemu, extrahovane denotace by jsme potom nechali potvrdit uzivateli. To by nam umozilo merit jak caste se trefime live, merit jaka je presnost ko konfirmaci, a rozsirit databazi. To by mel byt nas plan do conlusion (evaluation) jak se vyrovnat s temi malo daty co mame. Experimentem nasbirat vice, a pouzit to co jsme udelali. Realny test technologie co se vyvinula. Ne jenom overeni presnost.}

The next Section~\ref{sec:task} introduces the task of denotation extraction in more details. 
Several models for the denotation extraction from answer hints are proposed in Section~\ref{sec:denotation_extraction_algorithm}. 
Section~\ref{sec:evaluation} describes evaluation of the models on the QDD. 
A discussion of the results is provided in~Section~\ref{sec:discussion}.
Finally, we conclude the paper in Section~\ref{sec:conclusion}.

\begin{figure}
	\setlength{\belowcaptionskip}{-10pt}
    \setlength{\abovecaptionskip}{-10pt}
    \begin{center}
	    \includegraphics[scale=0.75]{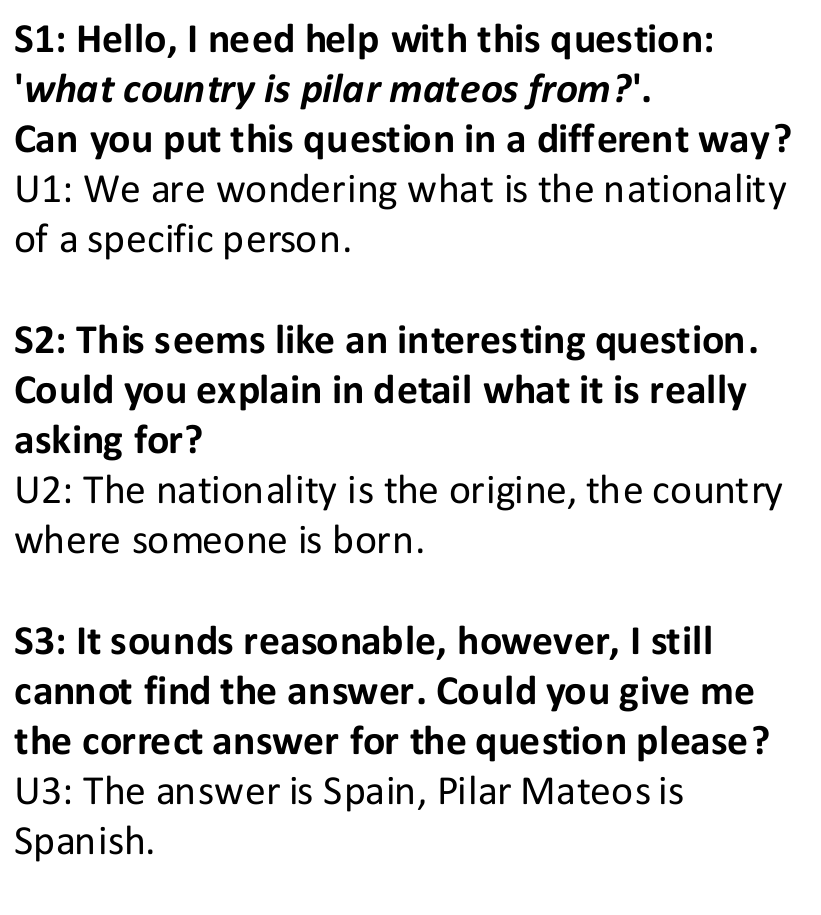}
	\end{center}
	
    \caption{An example of a short dialogue from the Question Dialogue Dataset containing multiple kinds of information. This work is only concerned with answer hints (U3).}
    \label{fig:example_dialogue}
\end{figure}

\section{The Task of Denotation Extraction} 
\label{sec:task}
In the framework of interactive learning by~\newcite{vodolan_rewochat2016}, an \emph{answer hint} is a piece of information obtained by asking a human user a question of the system's interest.
This way, one can get utterances as follows:
\dialogue{
  \system{What work did Scooter Libby write?}
  \user{Scooter Libby wrote a novel called The Apprentice.}
}

%In usual dialogue system setups, it is impractical to work directly with utterances in natural language. 
%, so it is usually processed into representation easier to handle by dialogue system's components. 
To make this information useful for a typical dialogue system, natural language utterances have to be mapped to a meaning representation.
In a question answering system, such a convenient meaning representation of an answer has a form of a denotation~\cite{Berant2013,liang2016learning,liang2016neural}.
A denotation is a set of entities, representing the correct answer of a question~\cite{Berant2013}, from the system's knowledge base (such as Freebase\footnote{https://developers.google.com/freebase/}). 
A knowledge base (KB) is a set of triplets in a form of \textit{subject entity}, \textit{relation}, \textit{object entity}.%\OD{Tuhle vetu muzes hodit do toho footnotu.}

The automatic mapping of answer hints to denotations is a challenging task. 
First, finding entities in an answer hint is difficult because a single entity may be described in many ways using natural language, and not all the ways will be captured in a KB. 
Second, the system has to select denotation entities from potentially several  entities in an answer hint.
Third, natural dialogues are prone to errors, speech disfluencies or misspellings (in text-based interfaces).
Finally, huge knowledge bases often contain lot of entities with the same label, but a different meaning (e.g., \textit{The Apprentice} is a novel, TV series, rock album, etc.).

The QDD (see Section~\ref{sec:introduction}) is a suitable testbed for the denotation extraction task. 
It is a set of dialogues between human users and a dialogue system where the system attempts to learn to communicate the content of it's KB to it's users.
%
%Dialogues in the QDD include several kinds of information that the dialogue system can use. 
%However, in this work we focus only on answer hints. 
%
As QDD contains manually annotated denotations, it can be used for the evaluation process. 
Note that the denotations for the QDD questions are single entities.%\OD{prohod predch. dve vety} \MV{prohozeni se mi nelibi - druha veta pocita s tim ze QDD ma anotovane denotace - proto prvni veta musi predchazet}
An example dialogue from the dataset can be seen in Figure~\ref{fig:example_dialogue}.

%There are also other datasets available for testing interactively learned systems, however, they often do not use a natural language~\cite{weston2016dialog}.

\section{Denotation Extraction Algorithm}
\label{sec:denotation_extraction_algorithm}
This section describes the process of denotation extraction. 
In this work, the denotation extraction algorithm is decomposed into two steps: an entity linking and a denotation identification, and it operates on pairs of a question and the corresponding answer hint.
% v prvni fazi vyhledame vsechny entity v answer hint utterance a odpovidajici otazce algoritmem popsanym v sekci [sekce]
%First, in the entity linking~\ref{ssec:entity_linking}, the answer hint's and the corresponding question's words are aligned to entities defined by a system's knowledge base.
First, entity linking jointly recognizes entities in a pair of a question and an answer hint and aligns these entities with a system's knowledge base (see Section~\ref{ssec:entity_linking}).
% ve druhe fazi mezi nalezenymi entitami identifikujeme entitu odpovidajici otazce [sekce]
%Second, in the denotation identification~\ref{ssec:denotation_identification}, denotation entities are identified among all entities found in the first phase.
Second, denotation identification selects the denotation from the answer hint's linked entities (see Section~\ref{ssec:denotation_identification}). 
% dosavadni vysledky dosazene na validacni testovaci sade z Question dialogues datasetu jsou shrnute v sekci [sekce]
%The results we got on QDD are summarized in Section~\ref{sec:evaluation}.
The evaluation of the algorithm is in Section~\ref{sec:evaluation}.

\subsection{Entity Linking}
\label{ssec:entity_linking}
% entity linking je intenzivne studovana disciplina dostupna formou internetovych sluzeb
%Entity linking is a task of identifying KB's entity mentions in a text~\cite{mcnamee2009}. 
The entity linking is a task of identifying entity mentions in a text~\cite{mcnamee2009} according to a system's KB. 
This task is intensively studied~\newcite{zhang2010entity,rao2013entity,pappu2017lightweight,hasibi2017entity}.
Also, many entity linking systems are available as web services\footnote{https://natural-language-understanding-demo.mybluemix.net/, https://azure.microsoft.com/cs-cz/services/cognitive-services/entity-linking-intelligence-service/ (links checked on 21.06.2017)}. 
One could consider using these services for this task. 
% z praktickych testu se vsak ukazuje ze bezne dostupne sluzby nejsou vhodne pro Question dialogues dataset
However, our informal experiments showed that these services do not perform well on the QDD.
% bezna slova totiz casto nebyvaji rozpoznana jako entita (male, female na Alchemy)
The reasons are twofold. 
First, some common entities in the QDD are not recognized (e.g. male, female, \ldots).
% tyto sluzby take nejsou odolne proti preklepum, ktere se v datech objevuji
Second, spelling errors appearing in the QDD are not handled gracefully.
%Manual inspection of the QDD pairs of a question and the answer hint suggest that many pairs are XXX. 
%\fj{Third, the services do not take advantage of the}
% z tohoto duvodu pouzivame pro entity linking vlastni algoritmus
Consequently, we propose a custom entity linking algorithm which handles the above problems.
% jeho princip je znazornen na obrazku [fig]

\begin{figure}	
    \setlength{\abovecaptionskip}{-5pt}
    \setlength{\belowcaptionskip}{-15pt}
    \begin{center}
	    \includegraphics[scale=0.55]{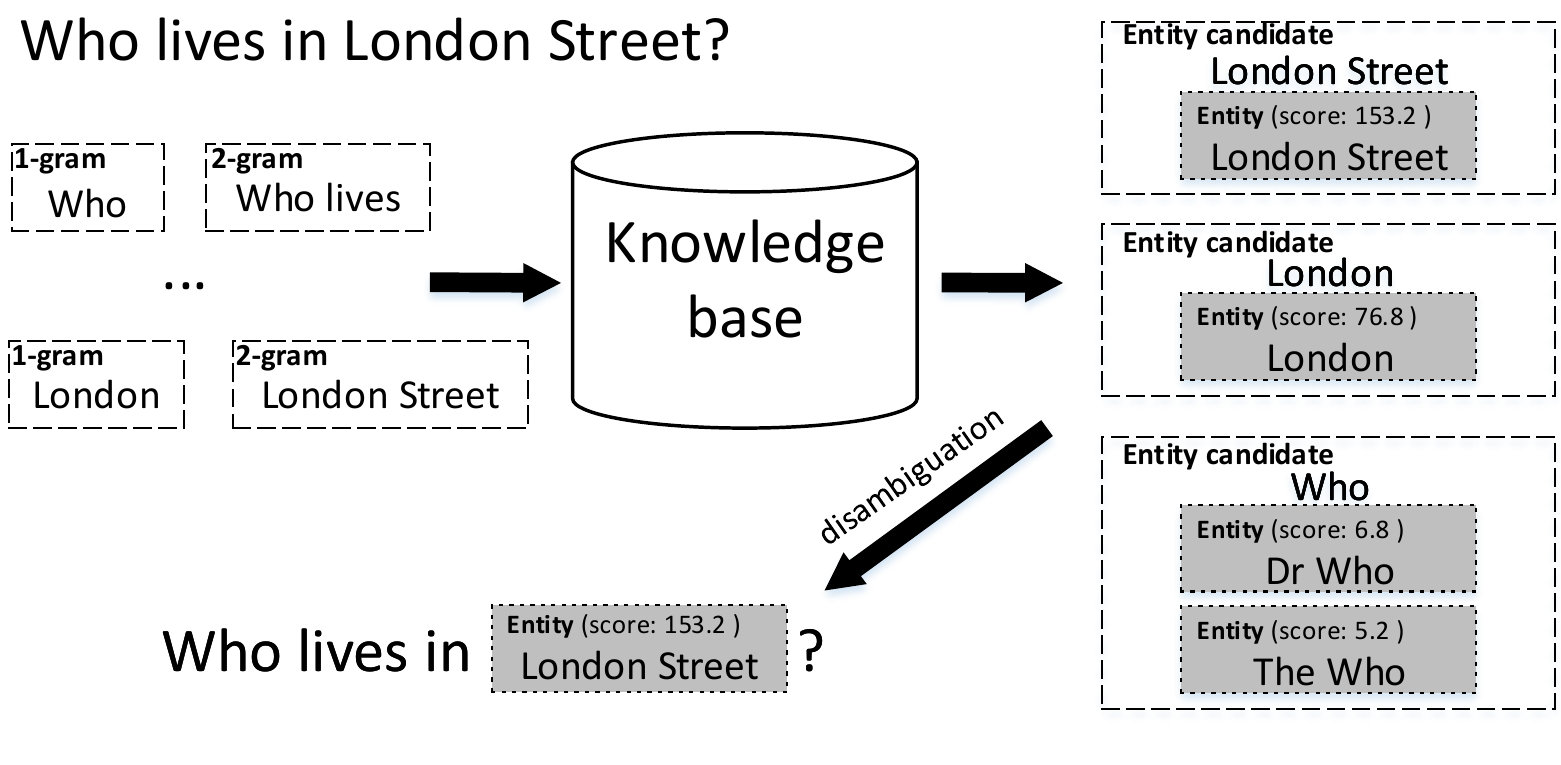}
	\end{center}
	
    \caption{Entity linking algorithm using similarity scores between n-grams and entities from knowledge base.}
    \label{fig:entity_linking}
\end{figure}

The principle of the proposed entity linking algorithm is shown in Figure~\ref{fig:entity_linking}.
% vstupem algoritmu je utterance v prirozenem jazyce a knowledge base systemu. 
%The algorithm takes an utterance in natural language and links utterance n-grams to corresponding entities in knowledge base.  \todo{mention context here somehow}
The algorithm takes a pair of a question and an answer hint as an input. 
First, the algorithm finds n-grams in both the question and the answer hint matching some entity name or its alias in the KB and marks the n-grams as entity candidates. 
The matching is done based on the string edit distance thus compensating for spelling errors. 
%in natural language and links utterance n-grams to corresponding entities in knowledge base.
%
%The algorithm works in the following steps:
%\setlist{nolistsep}
%\begin{enumerate}[noitemsep]
%  \item Identify spans of words as entity candidates. 
%  \item Link the entity candidates to entities.
%\end{enumerate}
% algoritmus nejprve najde pro kazdy ngram z linkovane vety ty entity, jejichz alias v knowledge-base se od ngramu prilis nelisi (mereno v editacni vzdalenosti, tak abychom dokazali kompenzovat preklepy)
%In the first step, algorithm finds n-grams matching some entity name according to knowledge base and marks them as entity candidates. 
%The matching is done based on edit distance, thus compensating for spelling errors.
%
% obvykle dostaneme nekolik kandidatu odpovidajicich prekryvajicim se ngramum (London, London Street)(ocekavame ze delsi nazvy jsou specifictejsi a proto presneji popisuji entitu)
%The algorithm chooses longer entity candidates in case they correspond to overlapping n-grams (e.g.,  \textit{London, London Street}) because we expect longer n-grams specify entities more precisely.
In case the entity candidates overlap (e.g.,  \textit{London, London Street}), the shorter candidates are discarded in favor of longer ones which presumably specify entities more precisely.
This heuristic has shown to be effective in our informal experiments.
Second, entity candidates are linked with entities in the KB.
For every entity candidate there are possibly many entities with the same matching name or alias while only one entity needs to be selected.
% disuambiguace, kterou vyuzivame vychazi z myslenky ze answer hinty nam davaji informace o vztazich mezi entitami, pri disambiguaci proto vybirame ty entity, ktere maji s ostatnimi entitami ve vete nejvice vztahu
For that purpose a \emph{relation maximization disambiguation} algorithm is used.

The algorithm is applied in two steps.
First, it links entities in a question. 
The algorithm selects the entities in a way that maximizes number of relations (according to the KB) between all the selected entities.
% pokud bychom nevyuzili informace z kontextu, nedokazali bychom rozlisit mezi xx kandidaty pro yyyyyy
Second, the algorithm links the entities in the corresponding answer hint. 
In addition to the answer hint entity candidates, the algorithm uses the linked entities from the question as a context.
In this case, answer hint entities are selected to maximize number of relations between each other and between them and the context entities.
This follows an intuition that relationship between an answer hint and its question can be expressed by relations of KB's entities. 
The following example shows how the relations can help to distinguish between entities corresponding to a single entity candidate:

\begin{figure}[H]	
	\includegraphics[scale=0.65]{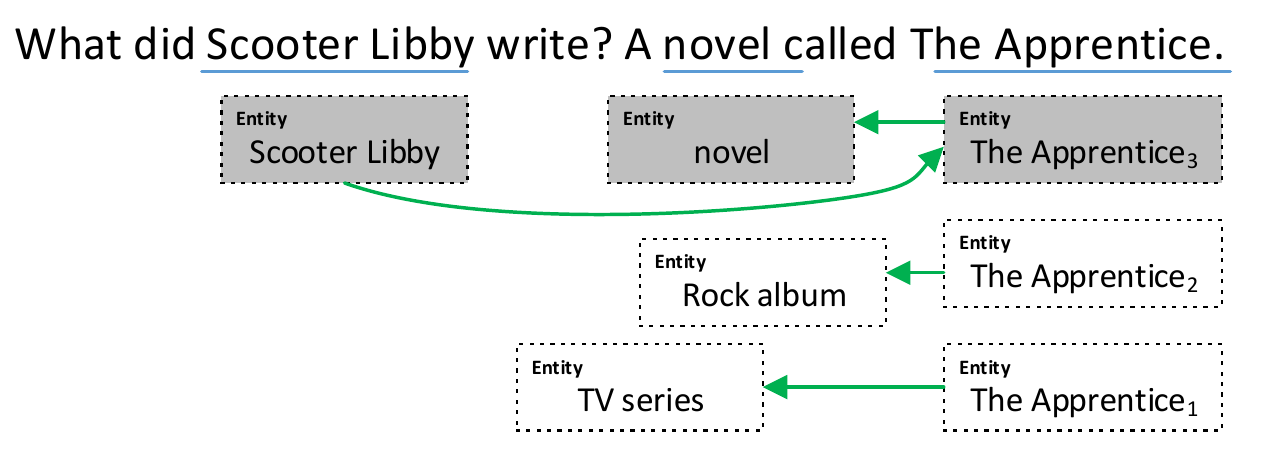}
\end{figure} \noindent
Entity candidate \textit{The Apprentice} matches three entities but only one is connected to other entities in the answer hint and question.
%and entities from a context.
%\fj{In the case of linking entities from answer hints, one can take advantage of the context,
%e.g. the question, by maximizing the number of relations of the answer hint entities together with the linked question entities.
%e.g. the question, by maximizing the number of relations between the answer hint entities and relations between the answer hint entities and the linked question entities. 
%For an answer hint we use a natural context -- the corresponding question.
%This follows an intuition that answer hints lexically represent information about the queried 
%entities and relations between them and the question.
%}
The result of this process is called a linked question and a linked answer hint.

As a baseline for the proposed relation maximization disambiguation algorithm, a popularity maximization disambiguation algorithm was also evaluated. In the popularity maximization, the entities for entity candidates are selected according to their so called \emph{popularity score} which is defined as a number of relations the entity has with all other entities in the KB, regardless the context.

%For results of the proposed algorithms on the QDD see~\ref{sec:evaluation}.

%\fj{This follows an intuition that answer hints entities are related to the question entities.}
%\fj{This follows an observation where the answer hint does not solely represent the denotation}

\subsection{Denotation identification}
\label{ssec:denotation_identification}
\begin{figure*}
	\setlength{\belowcaptionskip}{-10pt}
    \setlength{\abovecaptionskip}{-5pt}
    \begin{center}
	    \includegraphics[scale=0.85]{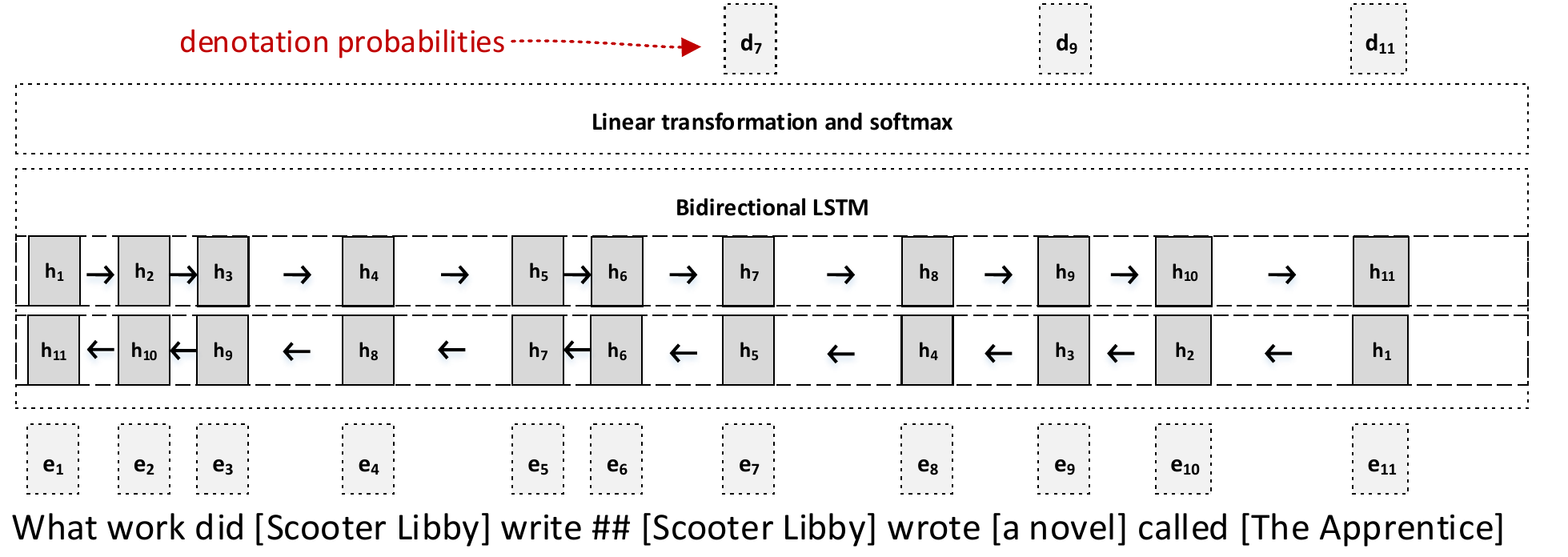}
	\end{center}
	
    \caption{The neural denotation identification model. An input sequence is made of question's and answer hint's words encoded as trainable embeddings $e_i$. The model outputs probabilities of being a denotation $d_i$.}
    \label{fig:bidir_model}
\end{figure*}

% ukolem teto faze je v answer hint (ziskany pomoci question rerouting [sekce]) nalezt entitu, ktera reprezentuje odpoved na otazku
The denotation identification algorithm selects a denotation among all answer hint entities detected during the entity linking (see Section~\ref{ssec:entity_linking}).
% obtiznost teto ulohy spociva v mnozstvi entit (answer kandidati), ktere se v answer hint vyskytuje (priklad)
The challenge of the denotation identification can be demonstrated on the following example (entities are enclosed by square brackets):%\OD{viz muj komentar v sekci \ref{sec:task}}
\dialogue{
  \system{What is [Sharon Calcraft]'s nationality?}
  \user{[Australian] [Composer] [Sharon Calcraft] was born in [1955] in [Sydney] [New South Wales] \textbf{[Australia]}.}
}
The answer hint in the example contains many entities, however, only \textbf{\textit{Australia}} entity is a denotation because it answers the corresponding question.

Two approaches for the denotation identification are proposed in this paper.
The first one, \textit{the context entity cancellation}, is a simple method based on an observation that most of non--denotation entities in the answer hint also appear in its corresponding question (see Section~\ref{ssec:simple_selection_models} for more details).
This model serves as a baseline for the later machine learned model.

The second approach, \textit{the attention selection model}, uses an attention-based bidirectional LSTM network, which is frequently used for identification of important parts of an input sequence~\cite{kadlec2016text,bahdanauCB14}.
This model and its variants are described in Section~\ref{ssec:attention_selection_models}.

\subsubsection{Simple Selection Models}
\label{ssec:simple_selection_models}

This section describes several variants of the context entity cancellation approach to the denotation identification.

\textbf{Basic cancellation} - The first, simplest approach is based on an observation that most of non--denotation entities within the answer hint come from its context -- the corresponding question. 
Therefore, the algorithm filters out all context entities from the answer hint entities.
From the remaining entities, the one with the highest Freebase popularity, which is a number of relations the entity has with other entities, is selected.

\textbf{+ enumeration detection} - The basic cancellation does not deal well with an enumeration in questions. See the following example:

\centeredutterance{\textbf{System:} Is Stana Katic male or female?}

\noindent The issue is that the question (the context) includes the correct answer which would be canceled by the basic cancellation model.
Therefore, the second algorithm uses an enumeration detection in questions (based on the keyword spotting). 
If enumeration is detected, the context entities are intersected with those in an answer hint instead of being subtracted as in the basic cancellation.

\textbf{+ context n-grams} - Next, the basic model cannot handle well a discrimination between a denotation and entities providing extra information commonly included in answer hints. 
See the following example: 
\dialogue{
	\system{Where was [Barack Obama] born?}
	\user{[Barack Obama] was a [USA president] born in [Hawaii].}
}
Even though, the user was asked about a place of birth, he/she also included information about a function of Barack Obama which is an additional information.
To deal with this issue, each entity popularity is multiplied by a prior probability of being a denotation (estimated from training data) given the surrounding context n-gram. 

In the example above, there are two entities left after the context cancellation: \textit{[USA president]} and \textit{[Hawaii]}. Examples of their corresponding 3-gram contexts are \textit{``was a \#ENTITY''}, \textit{``born in \#ENTITY''}. From the training data it is easy to count how many times those contexts appeared with \textit{\#ENTITY} being a denotation/being an extra entity, which is a sufficient information for computing the prior probability.

%according to surrounding n-grams, in the last baseline model.

\subsubsection{Attention Selection Models}
\label{ssec:attention_selection_models}
The other approach for the denotation identification uses an attention-based neural model over word sequences.
The word sequences are created by concatenation of a linked question and its linked answer, where each entity in a word sequence is encoded as a single word. 
See the example in Figure~\ref{fig:bidir_model}.
% vystupem modelu je pak pro kazdou entitu v sekvenci answer hintu pravdepodobnost, ze se jedna o answer entitu.
%The model outputs probabilities of being a denotation for every entity in answer hint's sequence.
For every answer hint entity, the model outputs a probability of being a denotation.

\textbf{Bidirectional attention model} - 
First, the model transforms the discrete word sequence into a vector representation using trained word embeddings. 
Then, a bidirectional LSTM layer converts word embeddings to context embeddings. 
Finally, the softmax layer produces a probability of being a denotation for every answer hint entity.
The architecture of the model can be seen in Figure~\ref{fig:bidir_model}. The minimized objective of the model is a categorical cross-entropy between the model's output and one-hot encoding of the denotation entity position (along the sequence length dimension).

\textbf{+ positional word features} - 
The next model variant, in addition, uses features about a position of an answer hint entity occurrence among the question entities.
These features are concatenated with word embeddings produced by the previous model and therefore must be generated for every word in the input sequence. 
One-hot encoding of the position with two special symbols is used.
If the entity does not appear among the question entities, it gets a NULL word symbol.   
If an input word is not an answer hint entity then it gets ZERO word symbol.
In Figure~\ref{fig:bidir_model} the first answer hint entity \textit{Scooter Libby} appears as first entity in the question and the second answer hint entity \textit{a novel} does not appear in the question at all.

%For example in Figure~\ref{fig:bidir_model}, the first answer hint entity appeared as the first entity in the question. 
%Second answer hint entity was not present in the question. 
%The next model variant, in addition, uses features about a position of every answer hint entity occurrence in a scope of question entities (e.g. first answer hint entity appeared as third entity in the question, second answer hint entity was not present in the question\dots). 

\textbf{+ pretrained glove} - To help the system perform better in setups with small amount of training data as is ours, influence of pretrained embeddings (for non-entity words) on model performance was explored.
The glove~\cite{pennington2014glove} embeddings were used due the simplicity of use in our framework.

\section{Evaluation}
\label{sec:evaluation}

% prezentovane modely jsme testovaly jsme testovali na datasetu QDD.
The proposed models were tested on the QDD which is divided into 950 training dialogues, 285 validation dialogues, and 665 test dialogues. 
These dialogues include both dialogues with correct answer hints and incorrect/incomplete answer hints.
In this work, only the dialogues with correct answer hints are used as it allows simple measurement of the denotation extraction performance.
Therefore, subsets of 176 training, 43 validation, and 132 test dialogues from the QDD were selected.

The entity linking models were implemented in C\# and optimized for the performance on large KBs. The surrounding context n-gram size was set to 3.

The denotation identification models were implemented in keras~\cite{chollet2015keras} and tensorflow~\cite{tensorflow2015-whitepaper}. They all used embedding size of 8, and 8 LSTM cells. The pretrained glove embeddings dimension was 10. 
The model parameters were optimized by Adam~\cite{kingma2014adam} with default hyper-parameters. The training ran for 50 epochs from which the best model parameters according to the validation accuracy were selected.

The evaluation considers three metrics: the entity linking accuracy, the denotation identification accuracy, and the denotation extraction accuracy.

The \textbf{entity linking accuracy} is measured as a ratio between answer hints containing a denotation (a QDD label) among their entities (i.e. correctly linked answer hints) and a count of all the answer hints. 
The results for the linking algorithm are shown in Table~\ref{tab:linking_results}. 
The results suggest that the entity linking with relation maximization outperforms the entity linking with popularity maximization. 
Manual inspection of errors shown the improvement comes from the use of  information about relations between linked entities. 

The QDD contains only a limited number of samples that can be used for testing. Therefore, the results have a binomial 95\% confidence intervals $\pm 0.09$, which is quite high. Narrowing of those intervals would require substantially more testing data which are not available yet.

\begin{table}[H]    
  \begin{center}
  \bgroup
  \def\arraystretch{1.05}
  \setlength{\tabcolsep}{3pt}
    \begin{tabular}{ll||c}
    & & accuracy \\[0.1cm] \hline 

    & relation maximization @1 & .628\\
    & relation maximization @5 & .628\\ \hline
    & popularity maximization @1 & .598\\ 
    & popularity maximization @2 & .621\\ 
    & popularity maximization @5 & .628\\ 

    \end{tabular}
  \egroup
  \end{center}
  \caption{Table with the accuracy of the entity linking. An answer hint is correctly linked when it contains a denotation. @n means that a correct answer was among n-best hypotheses.}
  \label{tab:linking_results}
\end{table}

Manual inspection of the errors shown that 
%apart from problems with ambiguity of the entity candidates 
the most of the denotations cannot be linked correctly as the wording of denotations significantly differ from names/aliases stored in the KB and a simple character string edit distance cannot account for these differences. See the following example:
%\vspace{1.5em}
\dialogue{
	\system{What is the nationality of [Steve Rassin]?}
	\user{[Steve Rassin] is American.}    
}
The entity [American] was not recognized although the database contains an entity with names like [United States of America], [USA], \ldots . 
In the context of interactive learning, it would be possible to learn the unknown aliases such as [American] directly from users by actively asking appropriate questions or from the context.

%\OD{bylo by pekny je nejak vic okomentovat -- jsou dobry?}
%\MV{nemam moc s cim srovnavat - random baseline bude 0, baseline na random disambiguation muze byt 1 ku 30 odhadem... tzn baseline by toho moc nerekla}

% Vykonnost modelu pro denotation identifikace ukazuje tabulka..
% merime ji pomoci dvou metrik. Prvni ukazuje s jakou presnosti model identifikuje entitu vzhledem ke vsem answer hintum v datasetu. 
%It is measured by two metrics. First, denotation extraction accuracy, shows how many denotations were correctly identified in a scope of all testset answer hints.
% druha pak pocita tuto presnost s vynechanim answer hintu, ve kterych ani jedna nalinkovana entita neodpovidala anotaci. 
%Second, denotation identification accuracy which takes only correctly linked answer hints into consideration. Therefore, it is a better quality measure of the denotation identification algorithm.

The \textbf{denotation identification accuracy} measures a ratio of the number of correctly identified denotations and the number of correctly linked answer hints. 
This measure evaluates quality of the denotation identification assuming a perfect linking algorithm.

The \textbf{denotation extraction accuracy} measures a ratio of the number of correctly identified denotations and the number of all answer hints. 
This is end-to-end measure for the denotation extraction - measure representing the overall chance of the proposed algorithms to identify correct denotation in an answer hint.
The results for the proposed denotation identification models are shown in Table~\ref{tab:answer_extraction_results}.
The results have a binomial 95\% confidence intervals $\pm 0.09$.
The results suggests that the attention based model with pretrained glove embeddings is comparable and possibly slightly better then the rule based baseline. 
This may be surprising as the training dataset consists of only 176 training examples.
While it is hard to come with firm conclusions given so little training and test data, the informal manual inspection of the test results suggests that the neural attention based model better disambiguates extra information entities from the denotations. 
It appears the neural model can learn the usage of prepositions before  denotations in the context of the corresponding question which the baseline system cannot.
%
%\fj{rychlost algorithmu, omezi plynouci z mnostvi entit v databazi?}
%\fj{qdd podmozina freebase? delka vety impact?}
%\fj{zhodnotit poresnost mereni. rict ze pocet prikladu neni moc a tak ty intervaly nejsou moc uzke. proc nename vice dat, kde je muzeme ziskat.}

\begin{table}[H]
\begin{center}
\bgroup
\def\arraystretch{1.05}
\setlength{\tabcolsep}{3pt}
\begin{tabular}{ll||c|c}
& & accuracy d.i.  & accuracy d.e. \\[0.1cm] \hline
 
& Basic cancellation  & .768& .477\\
& + enum. detection  & .768& .477\\ 
& + context ngrams  & .780& .485\\ \hline
& Bidir attention & .639 & .402\\ 
& + positional word features & .723 & .455\\ 
& + pretrained glove  & .793& .492\\ 

%& Basic cancellation  & .768$\pm$ .09& .477  $\pm$ .09\\
%& + enum. detection  & .768$\pm$ .09& .477  $\pm$ .09\\ 
%& + context ngrams  & .780$\pm$ .09& .485  $\pm$ .09\\ \hline
%& Bidir attention & .639$\pm$ .08 & .402 $\pm$ .08\\ 
%& + positional word features & .723$\pm$ .09 & .455 $\pm$ .09\\ 
%& + pretrained glove  & .793$\pm$ .09& .492 $\pm$ .09\\ 

\end{tabular}
\egroup
\end{center}
\caption{Table with denotation extraction accuracy and accuracy of denotation identification (when using our best linker). Notice that adding enum. detection did not improve test performance. However, it has shown to be useful on dev and train splits.}
\label{tab:answer_extraction_results}
\end{table}

The typical errors are caused by vague nature of some questions in the QDD. 
For example, in the case of "Who lives in the New York City?" there live millions of people in the city and user can answer with any subset of the New York citizens. However, QDD label contains only one of them, hurting the accuracy of the model.

The code for denotation identification is available at GitHub\footnote{
https://github.com/vodolan/DenotationIdentification}.

\section{Discussion}\label{sec:discussion}
The evaluation shows that denotations can be automatically extracted from the data obtained interactively from dialog system's users. 
The advantage of this approach compared to extraction from non-interactive sources is in ability of the system to immediately confirm its hypothesis (making sure the extraction was successful). This has shown as crucial in projects like NELL~\cite{mitchell2015nellwhuman}. Without humans in the loop, the NELL system was not able to automatically learn facts with a high accuracy.

The extracted denotations can be used for learning natural language understanding (NLU) components as shown in \newcite{Berant2013,liang2016learning,liang2016neural}. 
Therefore, dialogue system built around such NLU components, continuously trained on automatically extracted denotations, can adapt to long-term changes in language.

Another interesting field where denotations may be useful is a knowledge base population task~\cite{ji2011knowledge}. 
In this task, a KB is typically expanded by adding information extracted from off-line documents (e.g. Wikipedia pages).
However, this information can also be inferred from the denotations.
Therefore, the denotation extraction from answer hints allows to expand the KB by adding information obtained interactively from users.
The advantage of the later approach is in possibility to focus the extraction effort on entities of system's interest, thus making the process more efficient. 
In addition, the system can confirm extracted facts, ensuring the extracted information quality.

\section{Conclusion}\label{sec:conclusion}
We have presented a novel task which aims to support interactively learned dialogue systems by extracting denotations from natural dialogues with real users. 
We also proposed a method for solving the task and evaluated it on the Question Dialogues Dataset. 

The experiments shown that a reasonable amount of useful information can be extracted even with our simple rule-based baseline algorithms. 
Also, we have shown that it is possible to train a neural attention-based model, which slightly outperforms the baselines by using less than 200 training examples. 
%The code for denotation identification is available at GitHub.

In future work, we plan to extend the model to support multiple entity denotations to deal with multiple options answer hints corresponding to too broad (vague) questions. 
Further, we plan to devise methods for interactive learning of unknown aliases to entities in a KB to improve the linking accuracy.
%\fj{Finally and most importantly, we plan to run live deployment of the entity extraction algorithm in a system where each denotation would be confirmed with a user on-line. This deployment would allowed us to estimate  accuracy of the proposed solution after user based confirmations and also it would served as source of additional train/test data for this task.}

\section*{Acknowledgments}
This  work  was  funded  by  the  Ministry  of  Education,  Youth  and  Sports  of  the  Czech  Republic, SVV project 260 453, and GAUK grant 1170516 of Charles University in Prague. It used language resources stored and distributed by the LINDAT/CLARIN project of the Ministry of Education, Youth and Sports of the Czech Republic (project  LM2015071).

\bibliographystyle{IEEEbib}
\bibliography{paper}

\end{document}